\documentclass[runningheads]{llncs}
\usepackage[T1]{fontenc}
\usepackage[ruled,vlined,linesnumbered]{algorithm2e}
\usepackage{amsmath}
\usepackage{graphicx}
\usepackage{amssymb}
\usepackage{multirow, booktabs, makecell, colortbl}
\usepackage{xcolor}
\usepackage{listings}
\usepackage{caption} 
\usepackage{subcaption} 

\newcommand{\ourframework}{{\textsc{$\text{R}^2\text{R}$}}}

\newcommand{\ourframeworkfull}{{\text{Route-to-Rerank}}}
\newcommand{\twostgfull}{{\text{Entity Abstraction for Generalization}}}
\newcommand{\twostg}{{\textsc{EAG}}}

\definecolor{BlueGreen}{RGB}{0, 150, 150}
\newcommand{\gains}[1]{\textcolor{BlueGreen}{\tiny{($\uparrow$ #1)}}}   
\newcommand{\neutral}{\textcolor{gray}{\tiny{(=)}}}   
\definecolor{RedOrange}{RGB}{220, 50, 50} 
\newcommand{\loss}[1]{\textcolor{RedOrange}{\tiny{($\downarrow$ #1)}}} 

\begin{document}
\SetAlgorithmName{Procedure}{List of Procedures}{Procedure}
\title{\ourframework{}: A Route-to-Rerank Post-Training Framework for Multi-Domain Decoder-Only Rerankers}
\titlerunning{\ourframework{}: Post-Training Framework for Multi-Domain Decoder-Only Rerankers}
%
\author{
Xinyu Wang\inst{1,2}\thanks{Corresponding author: xinyu.wang5@mail.mcgill.ca} \and
Hanwei Wu\inst{1}$^{\dagger}$ \and
Qingchen Hu\inst{1,2}$^{\dagger}$ \and
Zhenghan Tai\inst{1,3}$^{\dagger}$ \\
Jingrui Tian\inst{1} \and
Lei Ding\inst{1,4}
\and 
Jijun Chi\inst{1} \and
Hailin He\inst{1} \and
Tung Sum Thomas Kwok\inst{1} \and
Yufei Cui\inst{2} \and
Sicheng Lyu\inst{1,2,5} \and
Muzhi Li\inst{6} \and
Mingze Li\inst{7} \and
Xinyue Yu\inst{1,7} \\
Ling Zhou\inst{8} \and
Peng Lu\inst{7}
}

\authorrunning{Wang et al.}

\institute{
\begin{tabular}{cccc}
\inst{1} SimpleWay.AI &
\inst{2} McGill University &
\inst{3} University of Toronto &
\inst{4} University of Manitoba \\
\inst{5} Mila &
\inst{6} CUHK &
\inst{7} Université de Montréal &
\inst{8} CG Matrix
\end{tabular}
\\[2mm]
$^{\dagger}$ Equal contribution
}

\maketitle              
\vspace{-0.5cm}
\begin{abstract}
Decoder-only rerankers are central to Retrieval-Augmented Generation (RAG). However, generalist models miss domain-specific nuances in high-stakes fields like finance and law, and naive fine-tuning causes surface-form overfitting and catastrophic forgetting. To address this challenge, we introduce \ourframeworkfull{} (\ourframework{}), a domain-aware framework that combines dynamic expert routing with a two-stage training strategy, Entity Abstraction for Generalization (EAG). EAG introduces a counter-shortcut mechanism by masking the most predictive surface cues (entities), forcing the reranker to learn domain-invariant relevance patterns rather than memorizing dataset-specific entities. To efficiently activate domain experts, we design a lightweight Latent Semantic Router that probes internal representations from the frozen backbone decoder of our reranker to select the optimal LoRA expert per query. 
Extensive experiments across different reranker backbones and diverse domains (legal, medical, and financial) demonstrate that \ourframework{} consistently surpasses generalist and single-domain fine-tuned baselines. Our results confirm that \ourframework{} is a model-agnostic and modular approach to domain specialization with strong cross-domain robustness.

\keywords{Retrieval-Augmented Generation \and Domain Adaptation \and Dynamic Routing \and LoRA \and Invariant Pattern Learning}
\end{abstract}

\section{Introduction}

The recent progress of generative Large Language Models (LLMs) has transformed NLP and enabled widespread real-world applications. However, despite their strong capabilities, LLMs still suffer from hallucination, brittle reasoning, and inconsistent knowledge recall. Retrieval-Augmented Generation (RAG) addresses these issues by grounding model outputs in external evidence. However, the reliability of a RAG system ultimately depends on its reranker, which selects the documents supplied to the generator~\cite{brown2025systematicliteraturereviewretrievalaugmented}. In high-stakes domains such as law and medicine,  accurate reranking is essential for trustworthy performance.

Decoder-only rerankers have become increasingly popular due to their strong semantic reasoning, inference efficiency, and compatibility with LLM-based retrievers~\cite{karpukhin2020densepassageretrievalopendomain}. However, most are trained as general-purpose models and struggle with domain-specific terminology, fine-grained intents, and long-tail knowledge. Their performance deteriorates under distribution shift~\cite{li2025lexragbenchmarkingretrievalaugmentedgeneration,li2023chatdoctormedicalchatmodel}, underscoring the need for reranking methods that remain robust in high-precision settings.

A common response to domain shift is fine-tuning on domain-specific data, but this approach often overfits to surface cues (e.g., company names, case IDs) and causes catastrophic forgetting of general ranking abilities. Evidence of this behavior is shown in Table~\ref{app:forgetting} in Appendix~\ref{app:forgetting}. The model adopts shortcut patterns rather than true relevance logic~\cite{kalajdzievski2024scalinglawsforgettingfinetuning,xiong2025oploraorthogonalprojectionlora}. Maintaining separate, fully fine-tuned models for each domain is also computationally impractical, and existing approaches—static adapters or heavy ensembles—struggle to balance specialization with efficiency~\cite{li2025ensembleslowrankexpertadapters,kong2024loraswitchboostingefficiencydynamic,zhang2025ragrouterlearningroutequeries}.


To bridge this gap, we propose \textbf{\ourframeworkfull{} (\ourframework{})}, a lightweight and modular framework for domain-adaptive reranking. \ourframework{} maintains a set of specialized experts implemented by LoRA adaptors~\cite{hu2021loralowrankadaptationlarge} and dynamically selects the appropriate expert for each query. Our two-stage training scheme, \textbf{\twostgfull{} (\twostg{})}, abstracts entity mentions to reduce shortcut learning and then fine-tunes on original data to enable specialization without forgetting. At inference time, a \textbf{Latent Semantic Router}, rather than an external classifier~\cite{10.1145/1571941.1571945,li-etal-2008-classifying}, probes the frozen decoder-only reranker backbone to identify domain signals and activate the optimal expert without relying on external classifiers. In summary, our main contributions are as follows:
\begin{enumerate}
    \item \textbf{Two-stage training with \twostg{}.} We design a data curation and training pipeline that masks surface entities prior to domain specialization, reducing overfitting and encouraging the model to learn domain-invariant patterns.
    
    \item \textbf{Latent Semantic Router.} We introduce a lightweight router that leverages the \emph{frozen} reranker backbone to dynamically activate LoRA experts, eliminating the need for additional feature extraction modules.
    
    \item \textbf{Model-Agnostic Effectiveness.} We demonstrate that \ourframework{} consistently improves performance across multiple domains and reranker architectures, including Qwen3-Reranker \cite{qwen3embedding} and BGE-Reranker \cite{li2023making}. These results highlight the generality and adaptability of our approach for decoder-only rerankers.
\end{enumerate}

\begin{figure}
    \centering
    \includegraphics[width = 0.8\linewidth]{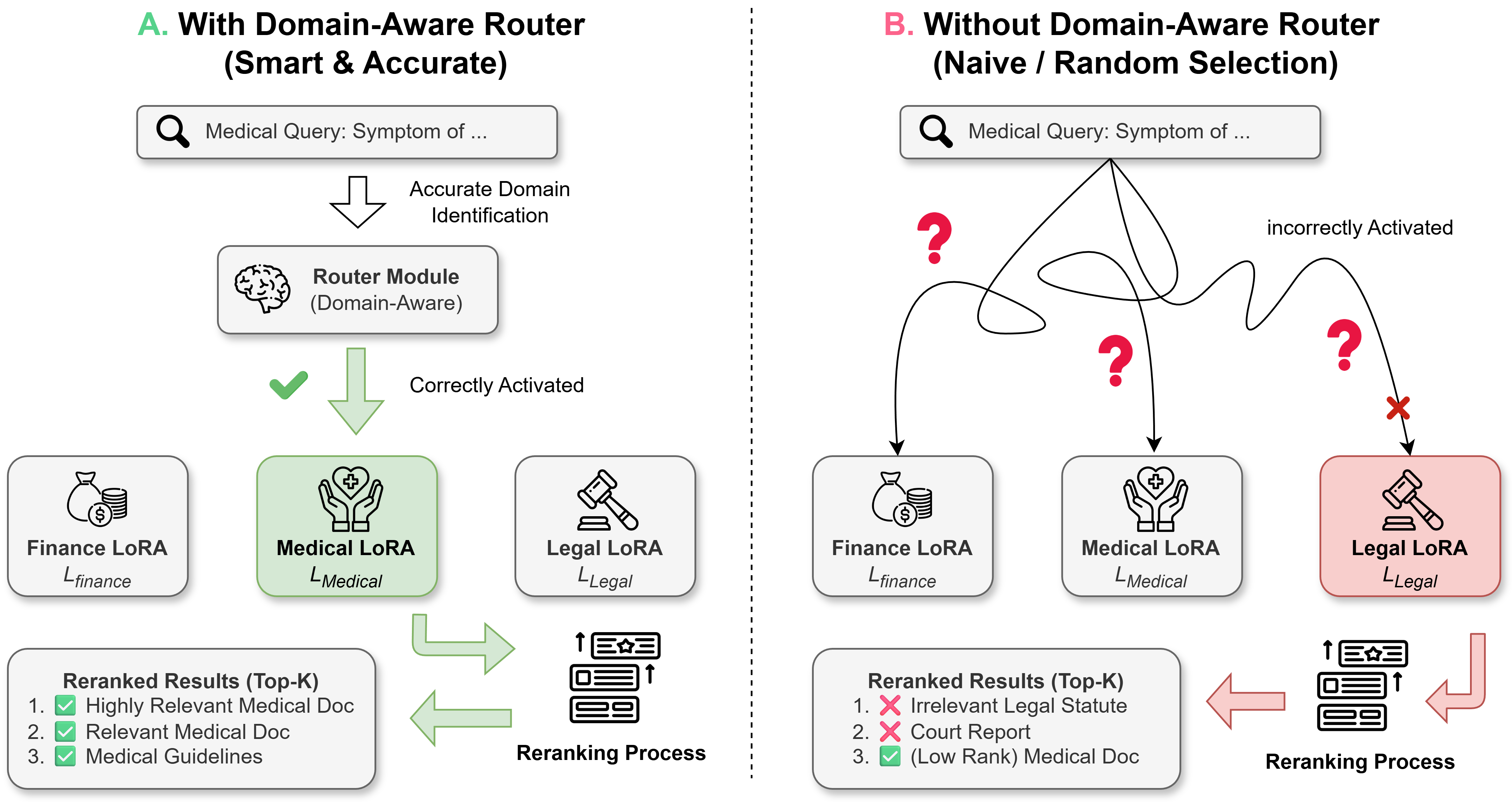}
    \caption{The impact of accurate domain routing on reranking quality. (A) A domain-aware router correctly activates a LoRA expert for a given query, maximizing in-domain expertise and precision. (B) Expert selection without proper routing results in domain mismatch and suboptimal reranking performance.}
    \vspace{-0.2cm}
    \label{fig:intro}
\end{figure}

\vspace{-0.5cm}
\section{Related Work}
\vspace{-0.3cm}
\subsection{Domain Adaptation and Parameter-Efficient Mining}
While RAG systems have improved LLM reliability, they remain vulnerable to domain distribution shifts. Benchmarks in high-stakes fields like law and medicine reveal that generalist rerankers struggle to distinguish fine-grained relevance signals amidst specialized terminology~\cite{li2025lexragbenchmarkingretrievalaugmentedgeneration,li2023chatdoctormedicalchatmodel}. 
To address this, Parameter-Efficient Fine-Tuning (PEFT) methods, e.g., LoRA ~\cite{hu2021loralowrankadaptationlarge}, have been adopted to inject domain knowledge without retraining the full backbone. However, naive PEFT often leads to overfitting on surface forms (e.g., specific names) or catastrophic forgetting of general capabilities ~\cite{kalajdzievski2024scalinglawsforgettingfinetuning,xiong2025oploraorthogonalprojectionlora}. In contrast, our work treats adaptation as a robust pattern mining task, employing adversarial EAG to force the model to learn invariant structural matching patterns.

\subsection{Dynamic Routing and Conditional Computation}
Dynamic computation, or the ability to conditionally activate network modules, offers a pathway to efficient multi-domain adaptation. This paradigm shares roots with Mixture-of-Experts (MoE) frameworks~\cite{chen2025lifelongknowledgeeditingvision,li2025ensembleslowrankexpertadapters,zhuang2025ldmolelearnabledynamicrouting}. In retrieval, recent works like RagRouter ~\cite{zhang2025ragrouterlearningroutequeries} and LoRA-Switch ~\cite{kong2024loraswitchboostingefficiencydynamic} try to route queries to different adapters, but they often rely on external classifiers or shallow embeddings that miss deeper semantic intent ~\cite{li-etal-2008-classifying}. In contrast, our $R^2R$ framework introduces a \textit{Latent Semantic Router} that inspects the frozen reranker’s internal representations to precisely activate the right LoRA expert without additional overhead.

\begin{figure*}[h] 
    \centering 
    \includegraphics[width = 0.85\linewidth]{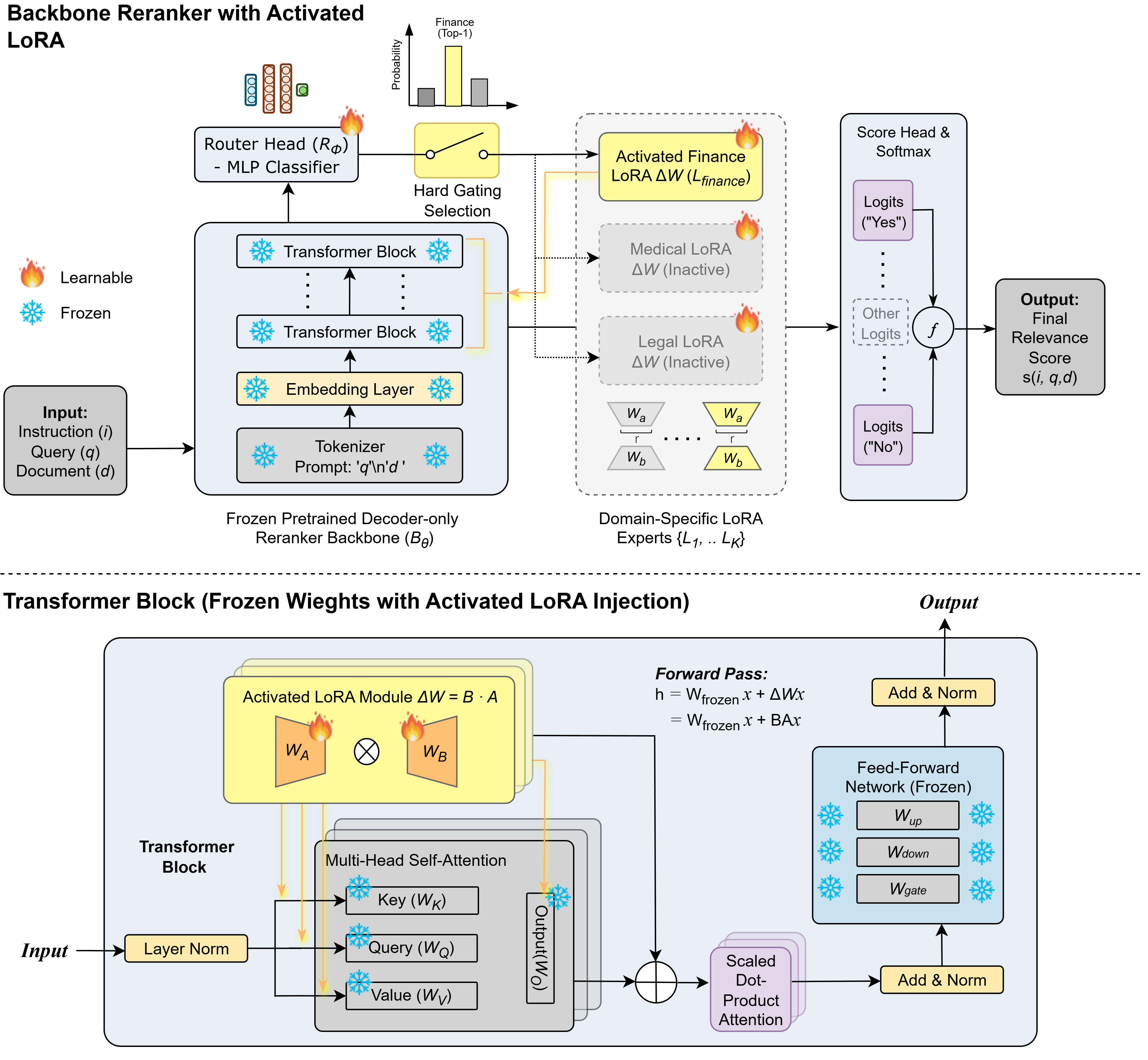}
\caption{Overview of the \ourframework{} framework. 
\textbf{Top:} The full Route-to-Rerank pipeline. The two-stage EAG curriculum first abstracts entities to learn invariant relevance patterns, then specializes on original domain data to produce domain-specific LoRA experts. During inference, the Latent Semantic Router probes the frozen backbone to select the appropriate expert. 
\textbf{Bottom:} The LoRA-augmented transformer block, where lightweight domain-specific LoRA adapters attach to the frozen reranker and are dynamically activated by the router.}

    \vspace{-0.2cm}
    \label{fig:overall_methodology}
\end{figure*}
\section{Preliminaries and Problem Formulation}
\vspace{-0.2cm}
\subsection{Generative Reranking Formulation}
\label{sec:decoder_reranker}
\vspace{-0.2cm}

We utilize a decoder-only LLM as the backbone for relevance estimation, formulating reranking as an instruction-aware next-token prediction problem.

\noindent\textbf{Input \& Architecture.} Given a query $q$, document $c$, and instruction $I$, we construct the input sequence $x$ via a template $\mathcal{T}$: $x = [\texttt{<Instruct>}: I; \texttt{<Query>}: q; \texttt{<Document>}: c]$.
The sequence is processed by $L$ transformer layers. For a hidden state $H^{(l)}$, the layer output is:
\begin{equation}
    \label{eq:transformer_block}
    H^{(l+1)} = \text{FFN}(\text{LN}(\tilde{H}^{(l)})) + \tilde{H}^{(l)}, \quad \text{where } \tilde{H}^{(l)} = \text{MHSA}(\text{LN}(H^{(l)})) + H^{(l)}.
\end{equation}
Here, $\text{MHSA}$ denotes Multi-Head Self-Attention utilizing the standard scaled dot-product mechanism.

\noindent\textbf{Relevance Quantification.} 
We extract the logit vector $z \in \mathbb{R}^V$ corresponding to the last token of $x$. Let $v_{\text{yes}}, v_{\text{no}}$ be the indices for tokens ``Yes'' and ``No''. The relevance score $s(q, c)$ is computed via a binary Softmax:
\begin{equation}
    s(q, c) = \frac{\exp(z_{\text{yes}})}{\exp(z_{\text{yes}}) + \exp(z_{\text{no}})}
= \sigma\left(z_{\text{yes}} - z_{\text{no}}\right).
\end{equation}
This maps the generative capability of the LLM to a discriminative ranking score $s \in (0,1)$, where $\sigma(\cdot)$ denotes the sigmoid function.

\subsection{Parameter-Efficient Adaptation (LoRA)}
\label{subsec:lora}
LoRA~\cite{hu2021loralowrankadaptationlarge} prevents catastrophic forgetting by freezing the pretrained weights $W_0 \in \mathbb{R}^{d \times k}$ and injecting trainable low-rank matrices $A \in \mathbb{R}^{r \times k}, B \in \mathbb{R}^{d \times r}$ ($r \ll d$). The forward pass is modified as:
\begin{equation}
    h = W_0 x + \frac{\alpha}{r} B A x,
\end{equation}
where $\alpha$ is a scaling hyperparameter. For notational simplicity, we omit the scaling factor $\alpha/r$ in subsequent sections and implicitly absorb it into the update term $\Delta W$. This modular design allows us to encapsulate domain-specific knowledge into lightweight expert modules $\Delta W_k$, serving as the basis for our routing framework.

\subsection{Problem Formulation}
\label{sec:problem_formulation}

Let $\mathcal{D} = \{d_{gen}\} \cup \{d_k\}_{k=1}^K$ be the domain set. The goal is to learn a scoring function $s_\theta(q, c)$ that ranks relevant candidates higher than negatives, formulated over training triplets $\tau = (q, c^+, \{c^-\})$.

Standard fine-tuning optimizes static parameters $\theta^*$ directly on target data. This approach suffers from \textbf{Shortcut Learning}, where models overfit surface forms rather than invariant structures (Appendix~\ref{app:forgetting}), and \textbf{Latent Ambiguity}, as the domain index $k$ is unobserved during inference.

To address this, we propose a \textbf{dynamic parameterization} $\theta(q) = \theta_{\text{base}} + \Delta\theta_{\phi(q)}$, where $\Delta\theta$ represents the trainable LoRA experts (defined as $\Delta W$ in Sec.~\ref{subsec:lora}) and $\phi(q)$ infers the latent domain. We formulate the training objective as a stepwise optimization, prioritizing global structural invariance before domain specialization:
\begin{equation}
    \min_{\Theta, \phi} \underbrace{\mathbb{E}_{\tau \sim \mathcal{P}_{\text{abstract}}} [\mathcal{L}_{rank}]}_{\text{Global Structural Invariance}} + \underbrace{\sum_{k=1}^K \mathbb{E}_{\tau \sim \mathcal{P}_{\text{target}}^{(k)}} [\mathcal{L}_{rank}]}_{\text{Domain Specialization}}.
\end{equation}
Here, the first term utilizes a global entity-abstracted distribution $\mathcal{P}_{\text{abstract}}$ to force the model to learn invariant patterns. The second term refines the model on distinct domain distributions $\mathcal{P}_{\text{target}}^{(k)}$ for precision. \ourframework{} approximates this joint objective sequentially: \textbf{\twostgfull{}} first optimizes the abstract term (Stage 1) to establish a robust structural foundation, followed by the target term (Stage 2) for domain injection (Section \ref{sec:eag}), while the \textbf{Latent Semantic Router} resolves the assignment $\phi(q) \rightarrow k$ (Section \ref{sec:router}).
\section{Methodology: \ourframeworkfull{} (\ourframework{})}
\label{sec:methodology}


Our proposed \textbf{\ourframeworkfull{} (\ourframework{})} method consists of two components: (1) a two-stage training strategy: \textbf{\twostgfull{}} (\textbf{\twostg{}}), and (2) a \textbf{Latent Semantic Router} for dynamic LoRA expert selection.

\subsection{Mining Invariant Patterns via \twostg{}}
\label{sec:eag}
\twostg{} mitigates overfitting to surface entities by progressively training the model through two stages. \textbf{Stage 1 (Counter-shortcut Entity Abstraction)} constructs an abstract dataset $D_\text{abstract}$ by replacing domain-specific named entities with randomized, type-consistent placeholders (e.g., "Zeekr" → \texttt{[COMPANY\_A]}). This encourages the model to learn structural relevance patterns rather than memorize specific names (i.e., taking "shortcuts"). By structural relevance patterns, we mean the relational structures among entities that indicate relevance, such as company–product links in finance, case–statute correspondences in law, or disease–symptom causal relations in medicine. After acquiring structural competence, \textbf{Stage 2 (Domain Specialization)} fine-tunes the model on the original, unmasked target dataset $D_\text{target}$, injecting precise domain knowledge while preserving general reasoning ability.
\vspace{-0.4cm}

\subsubsection{Automated Dataset Curation}
To support this pipeline, we employ a retriever-guided data curation process. For each query $q$, we construct a training triplet $(q, \mathcal{P}_q, \mathcal{N}_q)$, where the positve set  $\mathcal{P}_q$ is annotated by an LLM, while the negative set $\mathcal{N}_q$ is composed of both \textbf{Hard Negatives} (irrelevant chunks with high retrieval scores) and \textbf{Random Negatives}. This combination ensures the model learns to discriminate fine-grained semantic differences while maintaining broad separability. The detailed curation process is outlined in Algorithm~\ref{alg:EAG_curation}.
\begin{algorithm}[t]
    \SetAlFnt{\footnotesize}  
    \SetAlCapFnt{\footnotesize} 
    \SetAlCapNameFnt{\footnotesize} 
    \caption{Domain Dataset Curation Strategy}
    \label{alg:EAG_curation}
    \KwIn{Retriever $\mathcal{R}$; Target queries $Q_{\text{target}}$ and corpus $\mathcal{C}_{\text{target}}$.}
    \KwOut{Abstract domain dataset $D_{\text{abstract}}$ and specific datasets $\{D_{\text{target}}^{(k)}\}$.}

    $D_{\text{abstract}} \leftarrow \emptyset$\;
    \ForEach{target domain $k$}{
        $D_{\text{target}}^{(k)} \leftarrow \emptyset$\;
        \ForEach{query $q \in Q_\text{target}^{(k)}$}{
            $Candidates \leftarrow \mathcal{R}(q, \mathcal{C}_{\text{target}})$\;
            $(q, \mathcal{P}_q) \leftarrow \text{LLM\_Annotate}(q, Candidates)$\;
            
            $\mathcal{N}_q^{\text{hard}} \leftarrow Candidates \setminus \mathcal{P}_q$\;
            $\mathcal{N}_q^{\text{rand}} \leftarrow \mathrm{SampleRandom}(\mathcal{C}_{\text{target}} \setminus \mathcal{P}_q)$\;
            $\mathcal{N}_q \leftarrow \mathcal{N}_q^{\text{hard}} \cup \mathcal{N}_q^{\text{rand}}$\;
            
            Add $(q, \mathcal{P}_q, \mathcal{N}_q)$ to $D_{\text{target}}^{(k)}$\;
        }
        $D_{\text{abstract}} \leftarrow D_{\text{abstract}} \cup \text{ApplyAbstraction}(D_{\text{target}}^{(k)})$\;
    }
    \Return{$D_{\text{abstract}}$, $\{D_{\text{target}}^{(k)}\}$}\;
\end{algorithm}

\subsection{Optimization Objective}
\label{sec:optimization}

We train the LoRA experts using a contrastive learning objective. Given a query $q$, a positive chunk $c^+$, and a set of negatives $\{c^-_j\}_{j=1}^N$, the model computes relevance scores $s(q, c)$. The loss function minimizes the negative log-likelihood of the positive chunk:
\begin{equation} 
\label{eq:contrastive_loss}
\mathcal{L}_{\text{contrastive}} = -\log \frac{\exp(s(q, c^+)/\tau)}{\exp(s(q, c^+)/\tau) + \sum_{j=1}^{N} \exp(s(q, c^-_j)/\tau)},
\end{equation}
where $\tau$ is a temperature hyperparameter (set to 1.0 by default). This objective maximizes the margin between relevant and irrelevant evidence. The full two-stage training procedure is summarized in Algorithm~\ref{alg:EAG_training}, and training setups are detailed in Appendix~\ref{app:implementation}.

\begin{algorithm}[t]
    \SetAlFnt{\footnotesize}  
    \SetAlCapFnt{\footnotesize} 
    \SetAlCapNameFnt{\footnotesize} 
    \caption{Two-Stage EAG Fine-Tuning}
    \label{alg:EAG_training}
    \KwIn{Base model $\theta_{\text{base}}$; Abstract data $D_{\text{abstract}}$; Target data $D_{\text{target}}$.}
    \KwOut{Specialized LoRA parameters $\Delta\theta_{\text{expert}}$.}
    
    \SetKwFunction{Train}{ContrastiveTrain}
    \SetKwProg{Fn}{Function}{:}{}
    \Fn{\Train{$\theta, \mathcal{D}$}}{
        \While{not converged}{
            Sample batch $(q, c^+, \{c^-\})$ from $\mathcal{D}$\;
            Compute scores $s$ using Eq. (5)\; 
            Compute $\mathcal{L}_{\text{contrastive}}$\;
            Update $\theta$ via gradient descent\;
        }
        \Return{$\theta$}\;
    }
    
    \tcp{Stage 1: Learn Invariant Structure}
    $\theta_{\text{general}} \leftarrow$ \Train{$\theta_{\text{base}}, D_{\text{abstract}}$}\;
    
    \tcp{Stage 2: Inject Domain Knowledge}
    $\theta_{\text{expert}} \leftarrow$ \Train{$\theta_{\text{general}}, D_{\text{target}}$}\;
    
    \Return{$\theta_{\text{expert}}$}\;
\end{algorithm}
\vspace{-0.2cm}
\begin{algorithm}[t]
    \SetAlFnt{\footnotesize}  
    \SetAlCapFnt{\footnotesize} 
    \SetAlCapNameFnt{\footnotesize} 
    \caption{Router Training via Latent Probing}
    \label{alg:router}
    \KwIn{Query-Domain pairs $\{(q_i, d_i)\}$; Frozen backbone $f_{\theta}$; Router params $\phi = \{W_r, b_r\}$.}
    \KwOut{Trained router parameters $\phi$.}
    
    \ForEach{batch $(q, d)$}{
        $h_q \leftarrow \text{ExtractLastToken}(f_{\theta}(q))$\;
    
        $\hat{p} \leftarrow \text{softmax}(W_r h_q + b_r)$\;
    
        $\mathcal{L}_{\text{router}} \leftarrow -\sum d \log \hat{p}$\;
        Update $\phi$ to minimize $\mathcal{L}_{\text{router}}$\;
    }
    \Return $\phi$\;
\end{algorithm}

\subsection{Latent Semantic Router}
\vspace{-0.2cm}
\label{sec:router}

To enable dynamic expert selection during inference without incurring the latency of external classifiers, we introduce the \textbf{Latent Semantic Router}. 
Unlike traditional routing approaches that rely on shallow text embeddings, our router probes the \textbf{frozen backbone's internal world knowledge}.

Recall from Section~\ref{sec:decoder_reranker} that for any input query $q$, the reranker produces a final-token hidden state $h_q \in \mathbb{R}^d$. This vector $h_q$ contains a high-dimensional summary of the query's semantic intent. We project this representation through a lightweight routing head:
\begin{equation}
    p(d \mid q) = \text{softmax}(W_r h_q + b_r),
\end{equation}
where $W_r \in \mathbb{R}^{K \times d}$ and $b_r \in \mathbb{R}^K$ are the only trainable parameters for the router and $K$ denote the number of domains.

\textbf{Inference Mechanism.} During inference, the query is first passed through the frozen backbone. The router computes $p(d|q)$ and selects the domain expert $k^* = \arg\max_k p(d_k|q)$. The corresponding LoRA module $\Delta W_{k^*}$ is then dynamically activated to compute the final relevance score.

\textbf{Router Training.} The router is trained via standard cross-entropy loss on labeled query-domain pairs. Crucially, the backbone remains frozen, ensuring that the router learns to interpret the \textit{existing} semantic manifold of the LLM. The detailed training procedure is provided in Algorithm~\ref{alg:router}.

\vspace{-0.4cm}
\section{Experiments}
\vspace{-0.5cm}
\begin{figure}
    \centering
    \includegraphics[width=0.85\linewidth]{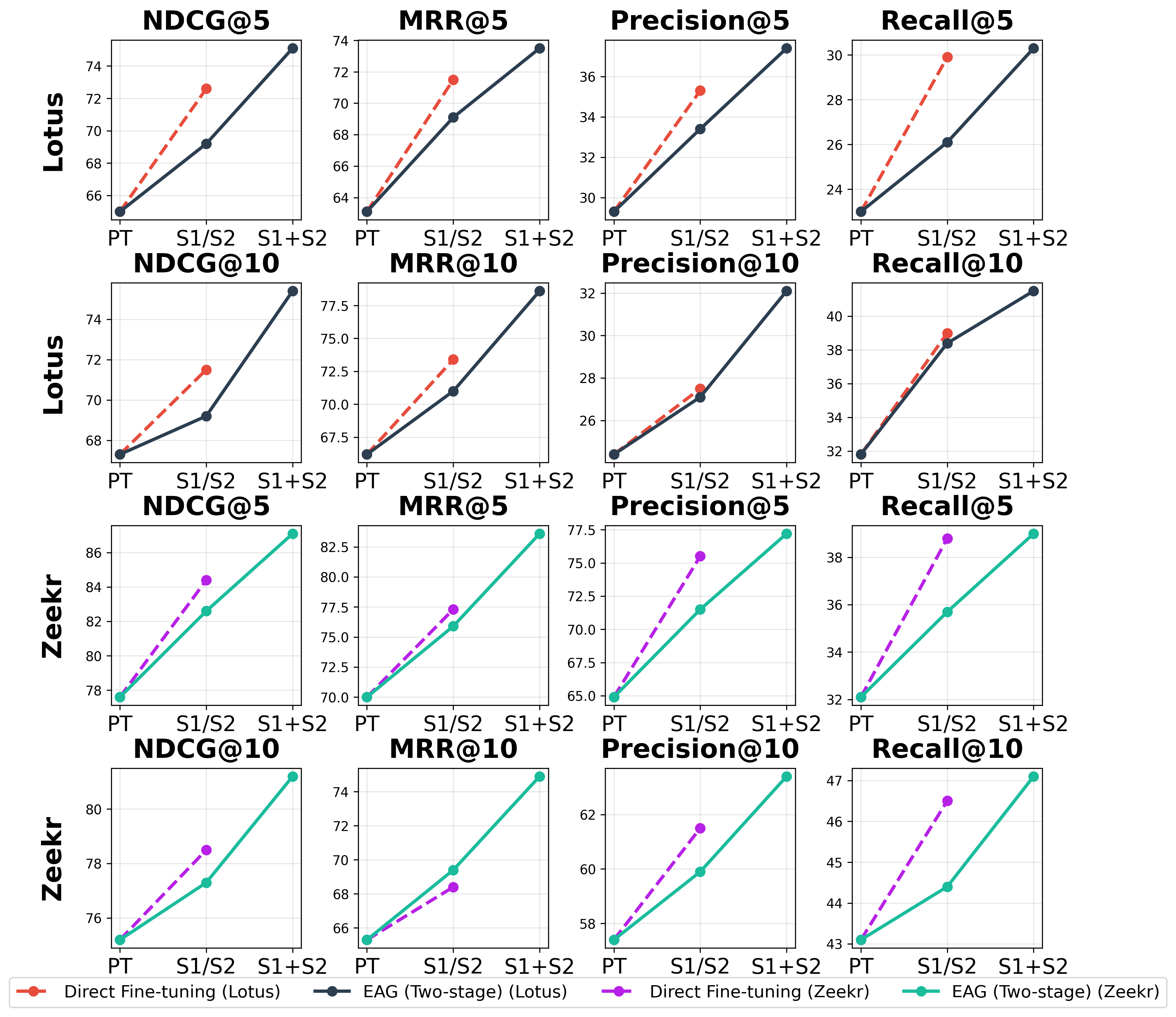}
    \caption{Reranker performance across training stages on Lotus and Zeekr datasets (PT=Pretrained, S1=Stage 1, S2=Stage 2). Dashed lines show direct fine-tuning (PT+S2), while solid lines show the two-stage EAG pipeline (PT+S1+S2). \twostg{} consistently outperforms direct fine-tuning across all metrics.}
    \label{fig:eag_eval}
\end{figure}
\vspace{-0.4cm}
\label{sec:experiments}
This section validates the superiority of the proposed \textbf{\ourframeworkfull{} ($\text{\ourframework{}}$)} framework and the \textbf{\twostg{}} training strategy across different models and datasets. Datasets and evaluation metrics are detailed in Appendix~\ref{app:datasets_metrics}.

\vspace{-0.7cm}
\subsection{Two-Stage \twostg{} Training Evaluation}
\vspace{-0.5cm}
\label{sec:eag_eval}
With setups detailed in Appendix \ref{app:two_stage_settings}, We evaluate whether the proposed two-stage \twostg{} pipeline improves reranking quality. Across both benchmarks and model variants, \twostg{} (PT+S1+S2) consistently outperforms direct fine-tuning (PT+S2), as shown in Table~\ref{tab:eag} and Figure~\ref{fig:eag_eval}. The Stage-1 abstraction step provides stable gains at both @5 and @10, confirming its effectiveness for domain specialization.

\vspace{-0.6cm}
\subsection{Router and End-to-End Evaluation}
\vspace{-0.5cm}
\label{sec:router_e2e_eval}
We evaluate routing quality and end-to-end reranking performance across the router configurations defined in Appendix~\ref{app:router_and_e2e_settings}. Table~\ref{tab:e2e_ablation_ranking} shows that our \textbf{Latent Semantic Router} has the highest routing quality. Combining \twostg{}-trained experts with our router, \ourframework{} achieves the strongest overall end-to-end results while maintaining the second lowest parameter overhead.
\vspace{-0.8cm}
\begin{table}[h]
\centering
\caption{Domain specialization results across two pretrained rerankers (PT=pretrained). For both datasets and both models, \twostg{} consistently provides the largest performance improvements over the pretrained baseline and outperforms direct fine-tuning.}
\label{tab:eag}
\small 
\setlength{\tabcolsep}{4pt} 

\resizebox{0.85\columnwidth}{!}{%
    \begin{tabular}{l | l | cccccc} %
    \toprule
    \multirow{2}{*}{\textbf{Dataset}}
    & \multirow{2}{*}{\textbf{Configuration}} 
    & \multicolumn{2}{c}{\textbf{NDCG}} 
    & \multicolumn{2}{c}{\textbf{MRR}} 
    & \multicolumn{2}{c}{\textbf{Recall}} \\
    \cmidrule(lr){3-4} \cmidrule(lr){5-6} \cmidrule(lr){7-8}
    & & @5 & @10 & @5 & @10 & @5 & @10 \\
    \midrule
    
    \multicolumn{8}{c}{\textbf{BAAI/bge-reranker-v2-gemma}} \\
    \midrule
    
    \multirow{3}{*}{\bfseries LexRAG}
    & PT & 81.1 & 82.4 & 78.6 & 79.0 & 90.6 & 91.6 \\
    & PT + direct FT & 90.4 \gains{9.3} & 90.6 \gains{8.2} & 89.7 \gains{11.1} & 89.6 \gains{10.6} & 93.9 \gains{3.3} & 94.3 \gains{2.7} \\
    & PT + \twostg{} & 92.5 \gains{11.4} & 92.6 \gains{10.2} & 92.0 \gains{13.4} & 92.0 \gains{13.0} & 93.9 \gains{3.3} & 94.7 \gains{3.1} \\
    \midrule
    
    \multirow{3}{*}{\bfseries ChatDoctor}
    & PT & 96.9 & 96.4 & 96.2 & 96.1 & 100.0 & 100.0 \\
    & PT + direct FT & 99.0 \gains{2.1} & 98.0 \gains{1.6} & 97.4 \gains{1.2} & 97.3 \gains{1.2} & 100.0 \neutral{} & 100.0 \neutral{} \\
    & PT + \twostg{} & 99.0 \gains{2.1} & 98.6 \gains{2.2} & 98.7 \gains{1.3} & 98.2 \gains{0.9} & 100.0 \neutral{} & 100.0 \neutral{} \\

    \midrule[\heavyrulewidth] 
    
    \multicolumn{8}{c}{\textbf{Qwen/Qwen3-Reranker-0.6B}} \\
    \midrule
    
    \multirow{3}{*}{\bfseries LexRAG}
    & PT & 86.8 & 87.5 & 85.7 & 85.9 & 92.4 & 94.4 \\
    & PT + direct FT & 91.3 \gains{4.5} & 90.1 \gains{3.4} & 90.3 \gains{4.6} & 91.2 \gains{5.3} & 93.2 \gains{0.8} & 94.6 \gains{0.2} \\
    & PT + \twostg{} & 95.8 \gains{9.0} & 95.9 \gains{8.4} & 94.9 \gains{9.2} & 96.0 \gains{10.1} & 96.7 \gains{4.3} & 96.3 \gains{1.9} \\
    \midrule
    
    \multirow{3}{*}{\bfseries ChatDoctor}
    & PT & 95.8 & 94.8 & 95.1 & 94.7 & 100.0 & 100.0 \\
    & PT + direct FT & 98.4 \gains{2.6} & 97.2 \gains{2.4} & 98.3 \gains{3.2} & 97.7 \gains{3.0} & 100.0 \neutral{} & 100.0 \neutral{} \\
    & PT + \twostg{} & 99.0 \gains{3.2} & 97.8 \gains{3.0} & 98.7 \gains{3.6} & 97.9 \gains{3.2} & 100.0 \neutral{} & 100.0 \neutral{} \\

    \bottomrule
    \end{tabular}
    
}
\end{table}
\vspace{-1.5cm}
\begin{table*}[h]
\small
\centering

\caption{Routing and end-to-end reranking results under different router configurations (LSR = Latent Semantic Router). The best score is shown in \textbf{bold} and the second best is \underline{underlined}. LSR attains the highest routing accuracy and macro F1, while \ourframework{} w/ LSR yields the strongest overall reranking performance with the second lowest parameter overheads.}
\label{tab:e2e_ablation_ranking}
\resizebox{\columnwidth}{!}{%
    \setlength{\tabcolsep}{4pt}
    \begin{tabular}{l | l | c c | *{3}{cc} | c}
        \toprule
        \multirow{2}{*}{\textbf{Configuration}} 
        & \textbf{Train}
        & \textbf{Router}
        & \textbf{Router}
        & \multicolumn{2}{c}{\textbf{NDCG}} 
        & \multicolumn{2}{c}{\textbf{MRR}} 
        & \multicolumn{2}{c}{\textbf{Recall}} 
        & \textbf{\# Extra}\\
        \cmidrule(lr){5-6} \cmidrule(lr){7-8} \cmidrule(lr){9-10}
        & \textbf{Strat.} & \textbf{Acc. (\%)} & \textbf{Macro F1} & \makebox[1.5em]{@5} & \makebox[1.5em]{@10} & \makebox[1.5em]{@5} & \makebox[1.5em]{@10} & \makebox[1.5em]{@5} & \makebox[1.5em]{@10} & \textbf{Params} \\
        \midrule
        1. Pretrained Reranker & None & N/A & N/A & 81.3 & 81.2 & 78.5 & 78.0 & 61.9 & 67.3 & \textbf{0} \\
        2. \ourframework{} w/ Sep. MLP Router & \twostg{} & 84.3 & 82.2 & 87.9 & 86.6 & 86.2 & 85.5 & 65.8 & 70.6 & 6.0B \\
        3. \ourframework{} w/ LLM as Router & \twostg{} & \underline{97.3} & \underline{97.3 }& \underline{88.8} & \underline{87.3} & \underline{87.2} & \underline{86.4} & \underline{66.2} & \underline{71.0} & 685B \\
        4. \textbf{\ourframework{} w/ LSR} & \textbf{\twostg{}} & \textbf{97.4} & \textbf{97.3} & \textbf{89.0} & \textbf{87.4} & \textbf{87.4} & \textbf{86.6} & \textbf{66.4} & \textbf{71.1} & \underline{0.2B} \\
        \bottomrule
    \end{tabular}
}

\end{table*}

\vspace{4cm}
\section{Conclusion}
\label{sec:conclusion}
\vspace{-0.3cm}
In this paper, we presented \ourframeworkfull{} (\ourframework{}), a lightweight post-training framework for domain-aware decoder-only rerankers. The method combines a backbone-probing router with the \twostgfull{} curriculum. This design helps the model specialize within each domain while staying robust across domains. Experiments across multiple domains and reranker backbones show clear in-domain gains over generalist baselines and simple fine-tuning, without sacrificing out-of-domain performance. Our analysis also shows that probing the frozen backbone with an LM-head classifier leads to much higher routing accuracy than a standalone MLP. Overall, \ourframework{} provides a practical and extensible approach for route-to-rerank domain adaptation in modern RAG systems.

\appendix
\section{Model Catastrophic Forgetting}
\vspace{-0.2cm}
\label{app:forgetting}
Table \ref{tab:forgetting} demonstrates that the model's general reranking capability degrades after fine-tuning. This observation motivates the need for parameter-efficient methods that can achieve specialization without compromising generalizability.

\begin{table}[!htbp]
    \centering
    \vspace{-0.4cm}
    \caption{Reranker (bge-reranker-v2-gemma) performance degradation on new domains after fine-tuning (4,000 steps) on the source domain (Zeekr and Lotus).}
    \label{tab:forgetting}
    \resizebox{0.8\columnwidth}{!}{%
    \begin{tabular}{l l c c c c c c}
        \toprule
        \textbf{Target} & \textbf{Reranker} & \multicolumn{2}{c}{\textbf{NDCG}} & \multicolumn{2}{c}{\textbf{MRR}} & \multicolumn{2}{c}{\textbf{Recall}} \\
        \cmidrule(lr){3-4} \cmidrule(lr){5-6} \cmidrule(lr){7-8}
        \textbf{Domain} & \textbf{Checkpoint} & \textbf{@5} & \textbf{@10} & \textbf{@5} & \textbf{@10} & \textbf{@5} & \textbf{@10} \\
        \midrule
        \multirow{2}{*}{\bfseries LexRAG}
        & Pretrained        & 81.1            & 82.4            & 78.6            & 79.0            & 90.6            & 91.6 \\
        & SFT (4,000 steps) & 77.7 \loss{3.4} & 79.2 \loss{3.2} & 74.0 \loss{4.6} & 74.6 \loss{4.4} & 89.2 \loss{1.4} & 88.8 \loss{2.8} \\
        \midrule
        \multirow{2}{*}{\bfseries ChatDoctor}
        & Pretrained        & 97.9            & 97.4            & 97.2            & 97.1            & 100.0           & 100.0 \\
        & SFT (4,000 steps) & 96.8 \loss{1.1} & 97.2 \loss{0.2} & 96.1 \loss{1.1} & 96.3 \loss{0.8} & 98.7 \loss{1.3} & 99.7 \loss{0.3} \\
        \bottomrule
    \end{tabular}}
    \vspace{-0.2cm}
\end{table}

\section{Model Training Setups}
\vspace{-0.2cm}
\label{app:implementation}
All reranker fine-tuning experiments use the same LoRA configuration across models. \textbf{Qwen3-Reranker-0.6B} is trained using the \texttt{Swift} \cite{zhao2024swiftascalablelightweightinfrastructure} framework, while \textbf{bge-reranker-v2-gemma} is trained using the 
\texttt{FlagEmbedding} \cite{bge_m3,bge_embedding} framework. Both rerankers are fine-tuned with the same LoRA configuration (rank~32, alpha~64, applied to the 
$q_{proj}$, $k_{proj}$, $v_{proj}$, and $o_{proj}$ layers).

\section{Experiment Setups}
\vspace{-0.2cm}
\label{app:exp_setups}
\subsection{Datasets and Evaluation Metrics}
\label{app:datasets_metrics}
We utilize four domain QA datasets to assess the domain adaptation capabilities of our framework: the \textbf{Legal Domain (LexRAG)} dataset \cite{li2025lexragbenchmarkingretrievalaugmentedgeneration}, which focuses on legal case retrieval and consultation; the \textbf{Medical Domain ($\text{ChatDoctor}$)} dataset \cite{li2023chatdoctormedicalchatmodel}, which consists of dialogues between patients and a specialized medical LLM, encompassing analysis of medical conditions and proposed treatment plans; and two subdomain datasets from the \textbf{Financial Domain (Zeekr and Lotus \cite{wang2025finsagemultiaspectragfinancial})}, focusing on retrieving information from financial filings.


We use standard information retrieval metrics at cutoffs $K=5$ and $K=10$: \textbf{NDCG@K}, \textbf{MRR@K}, \textbf{Precision@K}, and \textbf{Recall@K}; and we evaluate the quality of different routing mechanisms with \textbf{Accuracy} and \textbf{Macro F1 Score}. We omit $\text{Precision@K}$ for $\text{LexRAG}$ and $\text{ChatDoctor}$ since their queries correspond to a single ground truth chunk.

\subsection{\twostg{} Evaluation Settings}
\label{app:two_stage_settings}
We adopt a two-stage benchmarking process for \twostg{} evaluations. The performance baseline for Stage 1 is the pretrained base reranker. We select the checkpoint with the highest $\text{Precision@5}$ as the optimal Stage 1 model and use it as the base model for Stage 2 specialization. The baseline for Stage 2 is defined by a control experiment: the pretrained base reranker that is directly fine-tuned on the subdomain-specific dataset. This allows us to quantify the superiority of the two-stage \twostg{} approach over immediate subdomain specialization.

\subsection{Router and End-to-end Evaluation Settings}
\label{app:router_and_e2e_settings}
We use the same three routing configurations for both routing evaluation and end-to-end \ourframework{} experiments, all based on the Qwen3-Reranker-0.6B backbone and evaluated on the aggregated cross-domain dataset constructed from our selected domains. (1) \textbf{MLP Classifier} uses a standalone MLP fed by an external embedding model (bge-m3 in end-to-end settings), and its parameter cost includes both components. (2) \textbf{LLM-as-Router} sends the raw query to a general-purpose LLM (DeepSeek-V3), representing a high-capacity but API-dependent routing strategy. (3) \textbf{Latent Semantic Router} (ours) reuses the reranker's decoder to encode the query and adds only a lightweight MLP head on the last-token representation, requiring no external models.

These routing mechanisms are assembled into four end-to-end reranking variants: a vanilla pretrained reranker (no experts or routing); \ourframework{} with the MLP router; \ourframework{} with the LLM router; and \ourframework{} with our Latent Semantic Router. For all variants, we additionally report the total number of extra parameters introduced, counting both the routing module and all domain LoRA experts.


%
%
%
\bibliographystyle{splncs04}

\bibliography{citations, software}

\end{document}